\documentclass[letterpaper]{article} % DO NOT CHANGE THIS
\usepackage{aaai23}  % DO NOT CHANGE THIS
\usepackage{times}  % DO NOT CHANGE THIS
\usepackage{helvet}  % DO NOT CHANGE THIS
\usepackage{courier}  % DO NOT CHANGE THIS
\usepackage[hyphens]{url}  % DO NOT CHANGE THIS
\usepackage{graphicx} % DO NOT CHANGE THIS
\usepackage{natbib}  % DO NOT CHANGE THIS AND DO NOT ADD ANY OPTIONS TO IT
\usepackage{caption} % DO NOT CHANGE THIS AND DO NOT ADD ANY OPTIONS TO IT
\usepackage{algorithm}
\usepackage{algorithmic}

\usepackage{newfloat}
\usepackage{listings}
\usepackage{url}            % simple URL typesetting
\usepackage{booktabs}       % professional-quality tables
\usepackage{amsfonts}       % blackboard math symbols
\usepackage{nicefrac}       % compact symbols for 1/2, etc.
\usepackage{microtype}      % microtypography
\usepackage{xcolor}         % colors
\usepackage{amsmath,amssymb} % define this before the line numbering.
\usepackage{color}
\makeatletter
  \newcommand\figcaption{\def\@captype{figure}\caption}
  \newcommand\tabcaption{\def\@captype{table}\caption}
\makeatother
\usepackage{bm}
\usepackage{multicol,multirow}
\usepackage{algorithm,algorithmic}
\usepackage{graphicx}
\usepackage{subfig}
\usepackage{bbding}
\usepackage{colortbl}
\usepackage{arydshln}
\newenvironment{myproof}{\noindent{\bf Proof:}}{$\hfill \Box$ \vspace{10pt}}

\DeclareCaptionStyle{ruled}{labelfont=normalfont,labelsep=colon,strut=off} % DO NOT CHANGE THIS
\floatstyle{ruled}
\newfloat{listing}{tb}{lst}{}
\floatname{listing}{Listing}
\title{Resilient Binary Neural Network}
\author{
    Sheng Xu\textsuperscript{\rm 1}\equalcontrib,
    Yanjing Li\textsuperscript{\rm 1}\equalcontrib,
    Teli Ma\textsuperscript{\rm 2}\equalcontrib,
    Mingbao Lin,\\
    Hao Dong\textsuperscript{\rm 3},
    Baochang Zhang\textsuperscript{\rm 1,4}\thanks{Corresponding author: bczhang@buaa.edu.cn.},
    Peng Gao\textsuperscript{\rm 2},
    Jinhu L\"u\textsuperscript{\rm 1,4}
}
\affiliations{
    \textsuperscript{\rm 1} Beihang University, Beijing, P.R.China
    \textsuperscript{\rm 2} Shanghai AI Laboratory, Shanghai, P.R.China\\
    \textsuperscript{\rm 3} Peking University, Beijing, P.R.China
    \textsuperscript{\rm 4} Zhongguancun Laboratory, Beijing, P.R.China
}

\usepackage{bibentry}
\begin{document}
%\linenumbers
\maketitle
\begin{abstract}
Binary neural networks (BNNs) have received ever-increasing popularity for their great capability of reducing storage burden as well as quickening inference time. However, there is a severe performance drop compared with {real-valued} networks, due to its intrinsic frequent weight oscillation during training.
In this paper, we introduce a Resilient Binary Neural Network (ReBNN) to mitigate the frequent oscillation for better BNNs' training.
We identify that the weight oscillation mainly stems from the non-parametric scaling factor.
To address this issue, we propose to parameterize the scaling factor and introduce a weighted reconstruction loss to build an adaptive training objective. %To the best of our knowledge, it is the first work to solve BNNs based on a dynamically re-weighted loss function. 
%
%{\color{blue} We show  that the oscillation happens, if and only if the magnitude of update gradient     is greater than  the value of the coefficient parameter attached to the reconstruction loss.}
For the first time, we show that the weight oscillation  is  controlled by the balanced parameter attached to the reconstruction loss, which provides a theoretical foundation to  parameterize it in back propagation. %The oscillation only happens when the magnitude of gradient is big enough to change the sign of the latent weight.
Based on this, we learn our ReBNN by {calculating} the {balanced} parameter {based on} its maximum magnitude, which can  effectively mitigate the weight oscillation with a resilient training process.
Extensive experiments are conducted  upon various network models, such as ResNet and Faster-RCNN for computer vision, as well as BERT for natural language processing.
The results demonstrate the overwhelming performance of our ReBNN over prior arts. For example, our ReBNN achieves 66.9\% Top-1 accuracy with ResNet-18 backbone on the ImageNet dataset, surpassing existing state-of-the-arts by a significant margin. Our code is open-sourced at \url{https://github.com/SteveTsui/ReBNN}.

%a self-learning process, where the weighted factor of the introduced $\ell_2$ is updated based on its last-iteration gradient. which 
%of the efficient stabilizer to improve the learning of ReBNNs.
%We also give empirical analysis for our motivation and method.
%Our ReBNN marginally outperforms all prior state-of-the-art performance on various models and tasks in both CV and NLP areas {\em e.g.}, ResNet, Faster-RCNN and BERT, \textbf{with no extra parameters}. %The test models are available at \url{}.
\end{abstract}

\section{Introduction}

\label{sec1}
Deep neural networks (DNNs) have dominated the recent advances of artificial intelligence from computer vision (CV)~\cite{imagenet12,imagenet15} to natural language processing (NLP)~\cite{wang2018glue,qin2019stack} and many beyond.  In particular, large pre-trained models, \emph{e.g.}, ResNet~\cite{he2016deep} and BERT~\cite{devlin2018bert}, have continuously broken many records of the state-of-the-art performance. 
However, the achievement also comes with tremendous demands for memory and computation resources. 
These demands pose a huge challenge to the computing ability of many devices, especially resource-limited platforms such as mobile phones and electronic gadgets. 
In light of this, substantial research efforts are being invested in saving memory usage and computational power for an efficient online inference~\cite{he2018soft,rastegari2016xnor,qin2022bibert,li2022q}. 
%A number of model compression methods have been proposed, such as network pruning~\cite{lecun1990optimal,he2018soft,li2016pruning}, low-rank decomposition~\cite{denil2013predicting,lin2017espace}, network quantization~\cite{rastegari2016xnor,wang2018modulated,liu2018bi} , and knowledge distillation~\cite{romero2014fitnets} to accomplish this. 
%
Among these studies, network quantization is particularly suitable for model deployment on resource-limited platforms for its great reduction in parameter bit-width and practical speedups supported by general hardware devices.

\begin{figure*}[t]
    \centering
    \includegraphics[scale=.52]{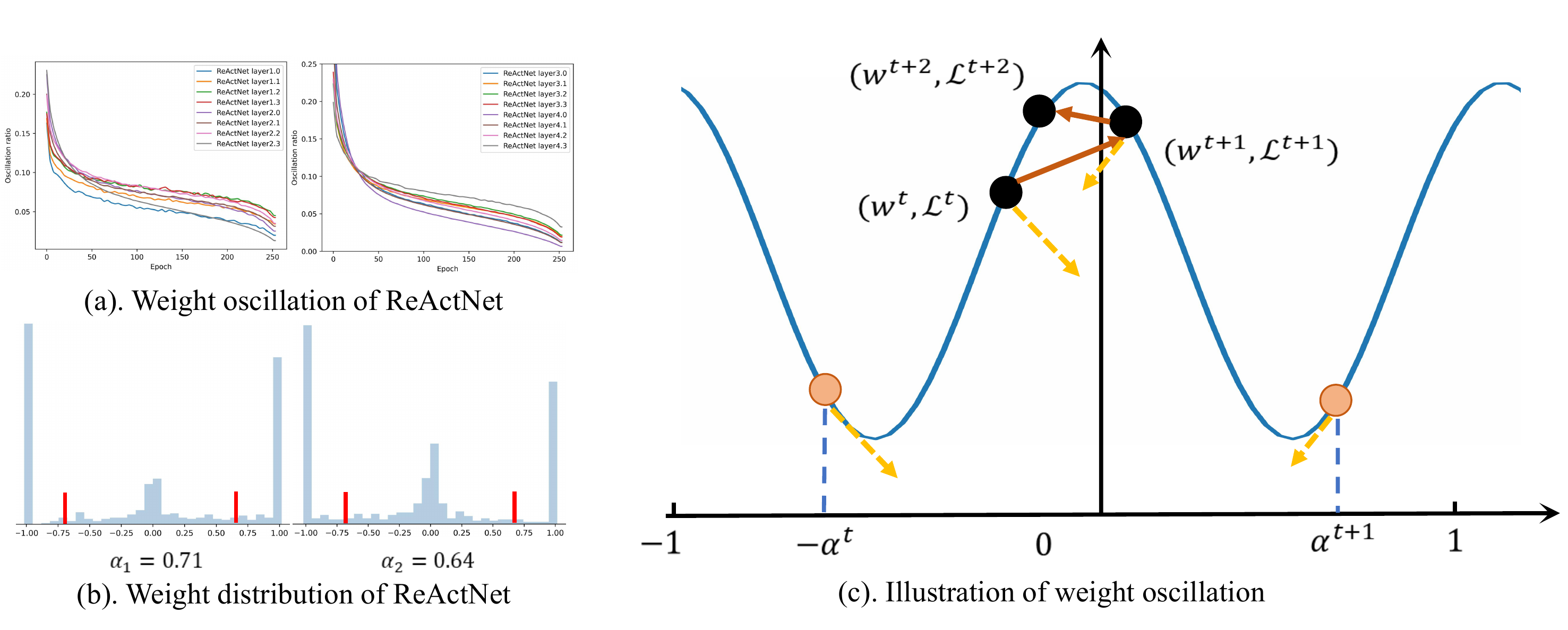}
    \caption{(a) We show the epoch-wise weight oscillation of ReActNet. (b) We randomly select 2 channels of the first 1-bit layer in ReActNet~\cite{liu2020reactnet}. Obviously, the distribution is with 3 peaks centering around $\{-1, 0, +1\}$, which magnifies the non-parametric scaling factor ({\color{red}\textbf{red}} line). (c) We
    illustrate the weight oscillation caused by such inappropriate scale calculation, where ${\bf w}$ and $\mathcal{L}$ indicate the latent weight and network loss function ({\color{blue}\textbf{blue}} line), respectively.}
    \label{fig:motivation}
\end{figure*}

Binarization, an extreme form of quantization, represents weights and activations of CNNs using a single bit, which well decreases the storage requirements by $32\times$ and computation cost by up to $58\times$~\cite{rastegari2016xnor}.
Consequently, binarized neural networks (BNNs) are widely deployed on various tasks such as image classification~\cite{rastegari2016xnor,liu2018bi,lin2022siman} and object detection~\cite{wang2020bidet,xu2021layer,xu2022ida}, and have the potential to be deployed directly on next-generation AI chips. However, the performance of BNNs remains largely inequal to the real-valued counterparts, due primarily to the degraded representation capability and trainability.

% Prior works~\cite{rastegari2016xnor,liu2020reactnet} prove that the scaling factor plays an essential role in the learning of BNNs. 
% However, most of prior works calculates the scaling factor in a non-parametric way using the channel-wise absolute mean absolute~\cite{liu2020reactnet}, which leads the weight oscillation and thus hindering the learning of BNNs.

Conventional BNNs~\cite{rastegari2016xnor,liu2020reactnet} are often sub-optimized, due to their intrinsic frequent weight oscillation during training. We first identify that the weight oscillation mainly stems from the non-parametric scaling factor. %
Fig.~\ref{fig:motivation}(a) shows the epoch-wise oscillation\footnote{A toy example of weight oscillation: From iteration $t$ to $t\!+\!1$, a misleading weight update occurs causing an oscillation from $-\!1$ to $1$, and from iteration $t$ to $t\!+\!2$ causes an oscillation from $1$ to $-\!1$.} of ReActNet, where weight oscillation exists even when the network is convergent. 
As shown in Fig.~\ref{fig:motivation}(b), the conventional ReActNet~\cite{liu2020reactnet} possesses a channel-wise tri-modal distribution in the 1-bit convolution layers, whose peaks respectively center around the $\{-1,0,+1\}$. Such distribution leads to a magnified scaling factor $\alpha$, and thus the quantized weights $\pm \alpha$ are quite larger than the small weights around $0$, which might cause the weight oscillation. As illustrated in Fig.~\ref{fig:motivation}(c), 
%Specifically, with the ${\rm sign}(\cdot)$ binarizer and straight-through-estimator (STE)~\cite{}, the gradient of small weights might be misleading and thus causing the weight oscillation and leading to less efficient training as well as a sub-optimal model. 
in BNNs, the real-valued latent tensor is binarized by the sign function and scaled by the scaling factor (the {\color{orange}\textbf{orange}} dot) in the forward propagation. 
In the backward propagation, the gradient is computed based on the quantized value $\pm \alpha$ (indicated by the {\color{yellow}\textbf{yellow}} dotted line). 
However, the gradient of small latent weights is misleading, when the scaling factor is magnified by the weights around $\pm 1$ as ReActNet (Fig.~\ref{fig:motivation}(a)). 
Then the update is conducted on the latent value (the \textbf{black} dot), which leads to the oscillation of the latent weight. With extremely limited representation states, such latent weights with small magnitudes frequently oscillate during the non-convex optimization.

To address the aforementioned problem, we aim to introduce a Resilient Binary Neural Network (ReBNN).
The intuition of our work is to re-learn the channel-wise scaling factor as well as the latent weights in a unified framework. 
Accordingly, we propose to parameterize the scaling factor {and introduce} a weighted reconstruction loss to build an adaptive training objective. 
We further show that the oscillation is factually controlled by the balanced parameter attached to the reconstruction loss, which provides a theoretical foundation to parameterize it in back propagation. 
The oscillation only happens when the gradient possesses a magnitude big enough to change the sign of the latent weight.
%Based on the new objective, we theoretically prove that such oscillation happens if and only if the latent weight gradient is greater than the weighted coefficient. 
Consequently, we calculate the balanced parameter based on the maximum magnitude of weight gradient during each iteration, leading to resilient gradients and effectively mitigating the weight oscillation.
%Based on the proof, we introduce a efficient self-stabilizer to improve the learning of ReBNNs. 
%We also analyze the effectiveness of the $\ell_2$ loss against the task-driven loss on the latent weight and scaling factor, based on which we introduce a gradient indicator to improve the convergence of BNNs. 
%We give empirical analysis and discussion about our motivation and metho.  
Our main contributions are summarized as follows:

\begin{itemize}
    \item We propose a new resilient gradient for learning the binary neural networks (ReBNN), which mitigates the frequent oscillation to better train BNNs. 
    
    \item We parameterize the scaling factor and introduce a weighted reconstruction loss to build an adaptive learning objective. We prove that the occurrence of oscillation is controlled by the balanced parameter attached to the reconstruction loss. Therefore, we utilize resilient weight gradients to learn our ReBNN and effectively mitigate the weight oscillation.
    %Based on the new back propagation process, we prove that ReBNN can theoretically reduce the weight oscillation. Based on the proof, we introduce a efficient self-stabilizer to improve the learning of ReBNNs. 
    
    \item Extensive experiments demonstrate the superiority of our ReBNN against other prior state-of-the-arts. For example, our ReBNN achieves 66.9\% Top-1 accuracy on the ImageNet dataset, surpassing prior ReActNet by 1.0\% with no extra parameters. In particular, our ReBNN also achieves state-of-the-art on fully binarized BERT models, demonstrating the generality of our ReBNN.
\end{itemize}

\section{Related Work}

BinaryNet, based on BinaryConnect, was proposed to train CNNs with binary weights.  The activations are triggered at run-time while the parameters are computed during training. Following this line of research, local binary convolution (LBC) layers are introduced in~\cite{juefei2017local} to binarize the non-linearly activations. XNOR-Net~\cite{rastegari2016xnor} is introduced to improve convolution efficiency by binarizing the  weights and inputs of the convolution kernels. More recently, Bi-Real Net \cite{liu2018bi} explores a new variant of residual structure to preserve the information of real activations before the sign function, with a tight approximation to the derivative of the non-differentiable sign function.
Real-to-binary~\cite{martinez2020training} re-scales the feature maps on the channels according to the input before binarized operations and adds a SE-Net~\cite{hu2018squeeze} like gating module. ReActNet ~\cite{liu2020reactnet} replaces the conventional PReLU~\cite{he2015delving} and the sign function of the BNNs with RPReLU and RSign with a learnable threshold, thus improving the performance of BNNs. RBONN~\cite{xu2022recurrent} introduces a recurrent bilinear optimization to address the asynchronous convergence problem for BNNs, which further improves the performance of BNNs.
However, most of these aforementioned suffer from the weight oscillation mainly stemming from the non-parametric scaling factor.

Unlike prior works, our ReBNN proposes to parameterize the scaling factor and introduces a weighted reconstruction loss to build an adaptive training objective. We further prove the oscillation is controlled by the balanced parameter. Based on the analysis, we introduce a resilient weight gradient to effectively address the oscillation problem.

\section{Methodology}
\subsection{Preliminaries}
\label{sec3.1}
Given an $N$-layer CNN model, we denote its weight set as ${\bf W} = \{{\bf w}^n\}_{n=1}^N$ and input feature map set as  ${\bf A} = \{{\bf a}^n_{in}\}_{n=1}^N$. The ${\bf w}^n \in \mathbb{R}^{C^n_{out}\times C^n_{in}\times K^n \times K^n}$ and ${\bf a}^n_{in}\in \mathbb{R}^{C^n_{in} \times W^n_{in} \times H^n_{in}}$ are the convolutional weight and the input feature map in the $n$-th layer, where $C^n_{in}$, $C^n_{out}$ and $K^n$ respectively stand for input channel number, output channel number and the kernel size. Also, $W^n_{in}$ and $H^n_{in}$ are the width and height of the feature maps.
Then, the convolutional outputs $\mathbf{a}^n_{out}$ can be technically formulated as:
\begin{equation}
{\bf a}^n_{out} = {\bf w}^n \otimes {\bf a}^n_{in},
\label{float_operation}
\end{equation}
where $\otimes$ represents the convolution operation. Herein, we omit the non-linear function for simplicity. 
Binary neural network intends to represent ${\bf w}^n$ and $ {\bf a}^n$ in a 1-bit format as 
${{\bf b}^{{\bf w}^n}}\in \{-1,+1\}^{C_{out}^n\times C_{in}^n \times K^n \times K^n}$ and ${{\bf b}^{{\bf a}^n_{in}}}\in\{-1,+1\}^{C^n_{in} \times W^n_{in} \times H^n_{in}}$ such that the float-point convolutional outputs can be approximated by the efficient XNOR and Bit-count instructions as:
\begin{equation}
{{\bf a}^n_{out}} \approx \bm{\alpha}^n \circ ({\bf b}^{{\bf w}^n} \odot {\bf b}^{{\bf a}_{in}^n})  ,
\label{binary_operation}
\end{equation}
where $\circ$ represents the channel-wise multiplication, $\odot$ denotes the XNOR and Bit-count instructions, and $\bm{\alpha}^n = \{ {\alpha}^n_1, {\alpha}^n_2, ..., {\alpha}^n_{C_{out}^n}\} \in \mathbb{R}^{C^n_{{out}}}_+$ is known as the channel-wise scaling factor vector~\cite{rastegari2016xnor} to mitigate the output gap between Eq.\,(\ref{float_operation}) and its approximation of Eq.\,(\ref{binary_operation}).
We denote $\mathcal{A} = \{ \bm{\alpha}^n \}_{n=1}^N$.
Most existing implementations simply follow earlier studies~\cite{rastegari2016xnor,liu2018bi} to optimize $\mathcal{A}$ and latent weights $\mathbf{W}$ based on a non-parametric bi-level optimization as:
\begin{align}
&{\bf W}^{*} = \mathop{\rm arg\;min}_{{\bf W}} \mathcal{L}({\bf W};\mathcal{A}^{*}),\\
{\rm s.t.} \;\;&\bm{\alpha}^{n*} = \mathop{\arg\min}_{{\bm{\alpha}}^n} \|{\bf w}^n-\bm{\alpha}^n\circ{\bf b}^{\mathbf{w}^n}\|_2^2,
\end{align}
where $\mathcal{L}(\cdot)$ represents the training loss.
Consequently, a closed-form solution of $\bm{\alpha}^n$ can be derived via the channel-wise absolute mean (CAM) as $\mathbf{\alpha}^n_i = \frac{\|{\mathbf{w}^n_{i,:,:,:}}\|_1}{M^n}$ and $M^n=C^n_{in}\times K^n \times K^n$.
For ease of representation, we use $\mathbf{w}^n_i$ as an alternative of $\mathbf{w}^n_{i,:,:,:}$ in what follows. 
The latent weight $\mathbf{w}^n$ is updated via a standard gradient back-propagation algorithm and its gradient is calculated as:
\begin{equation}\label{eq:backward0}
{{\delta }_{{\bf w}^n_i}}=\frac{\partial \mathcal{L}}{\partial {\bf w}^n_i}=\frac{\partial \mathcal{L}}{\partial \hat{{\bf w}}^n_i}\frac{\partial \hat{{\bf w}}^n_i}{\partial {{\bf w}^n_i}}
=\mathbf{\alpha}^n_i\frac{\partial \mathcal{L}}{\partial \hat{{\bf w}}^n_i} \circledast {\bf 1}_{|{\bf w}^n_i|\leq 1},
\end{equation}
where $\circledast$ denotes the Hadmard product and $\hat{{\bf w}}^n = \bm{\alpha}^n\circ{\bf b}^{{\bf w}^n}$.

\noindent\textbf{Discussion}.
Eq.\,(\ref{eq:backward0}) shows weight gradient mainly comes from the non-parametric $\mathbf{\alpha}^n_i$ and the gradient $\frac{\partial \mathcal{L}}{\partial \hat{{\bf w}}^n_i}$.
$\frac{\partial \mathcal{L}}{\partial \hat{{\bf w}}^n_i}$ is automatically solved in the back propagation and becomes smaller as network convergence, however, $\alpha^n_i$ is often magnified by the tri-modal distribution~\cite{liu2020reactnet}. Therefore, weight oscillation mainly stems from $\alpha^n_i$.
%
%\revise{where $\frac{\partial L}{\partial \hat{\bf w}}$ is automatically solved in the back propagation. Specifically, we consider the situation when the network approaches converging, the weight gradient of $\frac{\partial L}{\partial \hat{\bf w}}$ becomes smaller, while $\alpha^n_i$ is often magnified by the tri-modal distribution~\cite{liu2020reactnet} and appears to be more important issue for weight oscillation.} 
% As revealed in the introduction, $\alpha^n_i$ is often magnified by the tri-modal distribution~\cite{liu2020reactnet}. 
Given a single weight ${\bf w}^n_{i,j} (1\leq j \leq M^n)$ centering around zero, the gradient $\frac{\partial \mathcal{L}}{\partial {\bf w}^n_{i,j}}$ is misleading
, due to the significant gap between ${\bf w}^n_{i,j}$ and ${\alpha}^n_i{\bf b}^{{\bf w}^n_{i,j}}$. 
Consequently, the bi-level optimization leads to frequent weight oscillation. 
To address this issue, we reformulate traditional bi-level optimization using Lagrange multiplier and show that a learnable scaling factor is a natural training stabilizer.
% Given a weight centering around zero ${\bf w}_{i,j}$, the gradient $\frac{\partial \mathcal{L}}{\partial \hat{{\bf w}_i}}$ is probably misleading due to the significant gap between ${\bf w}_{i,j}$ and $\mathbf{\alpha}_i{\bf b}^{{\bf w}_{i,j}}$. 

% Moreover, due to $\mathbf{\alpha}_i$ is magnified by the tri-modal distribution, the ${\delta }_{{\bf w}_{i,j}}$ is misleading and magnified, leading to frequent weight oscillation.
%where such bi-level optimization 
%It motivates us to parameterize in BNNs to learning representative scaling factors and improve such distribution for efficient training.

\subsection{Resilient Binary Neural Network}
We first give the learning objective in this paper as:
\begin{equation}
\mathop{\arg\min}_{{\bf W}, \mathcal{A}} \mathcal{L}({\bf W}, \mathcal{A})+ \mathcal{L}_R({\bf W}, \mathcal{A}),
\label{11}
\end{equation}
where $\mathcal{L}_R(\mathbf{W}, \mathcal{A})$ is a weighted reconstruction loss and defined as:
\begin{equation}
\mathcal{L}_R({\bf W},\mathcal{A}) = \frac{1}{2}\sum^{N}_{n=1}\sum^{C_{out}}_{i=1}\gamma^n_{i}\|{{\bf w}^n_i}-{\mathbf{\alpha}}^n_i {\bf b}^{{\bf w}^n_i}\|_2^2,
\label{10}
\end{equation}
in which $\gamma^n_i$ is a balanced parameter. Based on the objective, the weight gradient in Eq.\,(\ref{eq:backward0}) becomes:
\begin{equation}
\begin{aligned}
    \delta_{{\bf w}^n_i} &= \frac{\partial \mathcal{L}}{\partial {\bf w}^n_i} + \gamma^n_i ({\bf w}^n_i - \mathbf{\alpha}^n_i {\bf b}^{{\bf w}^n_i}) \\
    &= \mathbf{\alpha}^n_i (\frac{\partial \mathcal{L}}{\partial \hat{\bf w}^n_i}\circledast {\bf 1}_{|{\bf w}^n_i|\leq 1} - \gamma^n_i {\bf b}^{{\bf w}^n_i}) + \gamma^n_i {\bf w}^n_i.
 \end{aligned}
 \label{eq:gradient}
\end{equation}

{\color{black} The $\mathcal{S}^n_i(\mathbf{\alpha}^n_i, {{\bf w}^n_i})= \gamma^n_i ({\bf w}^n_i - \mathbf{\alpha}^n_i {\bf b}^{{\bf w}^n_i})$ is an additional term added in the back-propagation process.
%
%Note that Eq.\,(\ref{eq:gradient}) avoids the constant weight due to the item $\gamma^n_i {\bf w}^n_i$.
%
We add this item given that a too small $\alpha_i^n$ diminishes the gradient $\delta_{{\bf w}^n_i}$ and causes a constant weight ${\bf w}_i^n$.
In what follows, we state and prove the proposition that $\delta_{{\bf w}^n_{i,j}}$ is a resilient gradient for a single weight ${\bf w}^n_{i,j}$.}
We sometimes omit subscript $i,j$ and superscript $n$ for an easy representation. 
% we denote $\mathcal{S}(\alpha, {\bf w}) = {\bf w} - \alpha {\bf b}^{\bf w}$ for simplicity. We omit subscript $i$ for simplicity.
% \begin{equation}
%     \begin{aligned}
%     \mathcal{S}(\alpha, {\bf w}) &= {\bf w} - \alpha {\bf b}^{\bf w} \\\frac{\partial \mathcal{L}}{\partial {\bf w}} &= \frac{\partial \mathcal{L}}{\partial {\bf w}} + \gamma \mathcal{S}(\alpha, {\bf w}) \\
%     &= \alpha (\frac{\partial \mathcal{L}}{\partial \hat{\bf w}} - \gamma {\bf b}^{\bf w}) + \gamma {\bf w}
%     \end{aligned}
% \end{equation}
%When the gradient directions of ${\bf w}$ and $\mathbf{\alpha}$ are conflicting, the back propagation of ${\bf w}$ are modified through $\mathcal{S}(\mathbf{\alpha}, {\bf w})$. Otherwise, the back propagation degenerate to the traditional weight updating. 

\noindent{\bf Proposition 1.}
\textit{The additional term $\mathcal{S}(\alpha, {\bf w})= \gamma ({\bf w} - \alpha {\bf b}^{{\bf w}})$ achieves a resilient training process by  suppressing frequent weight  oscillation. {Its balanced factor $\gamma$ can be considered as the parameter  controlling the occurrence of the weight oscillation.}}%\textit{The additional term $\mathcal{S}(\alpha, {\bf w})={\bf w} - \alpha {\bf b}^{{\bf w}}$ leads to a resilient training process by suppressing frequent weight  oscillation.}
%and weight gradient inaccurate effectively compared with STE gradients

\begin{myproof}
We prove the proposition by contradiction. 
For a single weight ${\bf w}$ centering around zero, the straight-through-estimator ${\bf 1}_{|{\bf w}|\leq 1} = 1$. Thus we omit it in the following.
Based on Eq.~(\ref{eq:gradient}), with a learning rate $\eta$, the weight updating process is formulated as:
\begin{equation}\label{eq:chain_rule}
    \begin{aligned}
    {\bf w}^{t+1} &= {\bf w}^{t} - \eta \delta_{{\bf w}^{t}} \\
    &= {\bf w}^{t} - \eta[\alpha^t(\frac{\partial \mathcal{L}}{\partial \hat{{\bf w}}^t} - \gamma {\bf b}^{{\bf w}^t}) + \gamma{\bf w}^{t}]\\
    &= (1 - \eta \gamma){\bf w}^{t} - \eta \alpha^t (\frac{\partial \mathcal{L}}{\partial \hat{\bf w}^{t}} - \gamma {\bf b}^{{\bf w}^{t}}) \\
    &= (1 - \eta \gamma)\big[{\bf w}^{t} - \frac{\eta \alpha^t}{(1 - \eta \gamma)}(\frac{\partial \mathcal{L}}{\partial \hat{\bf w}^{t}} - \gamma {\bf b}^{{\bf w}^{t}})\big],
    %&= (1 - \eta \gamma)[{\bf w}^{t} - \gamma_{\eta}(\frac{\partial \mathcal{L}}{\partial \hat{\bf w}} - \gamma {\bf b}^{\bf w})],
    \end{aligned}
\end{equation}
where $t$ denotes the $t$-th training iteration and $\eta$ represents learning rate. 
%
%We propose that {\color{red}${\bf w}$ with different distances from $\pm 1$}, the gradient should be modified according to their scaling factor and current learning rate. 
%
Different weights shares different distances to the quantization level $\pm 1$, therefore, their gradients should be modified in compliance with their scaling factors and current learning rate.
We first assume the initial state ${\bf b}^{{\bf w}^t} = -1$, and the analysis process is applicable to the case of initial state ${\bf b}^{{\bf w}^t} = 1$. 
The oscillation probability from the iteration $t$ to $t+1$ is:
\begin{equation}
    \begin{aligned}
    \left.P({\bf b}^{{\bf w}^t} \neq {\bf b}^{{\bf w}^{t+1}})\right|_{{\bf b}^{{\bf w}^t} = -1} \leq P(\frac{\partial \mathcal{L}}{\partial \hat{\bf w}^t} \le - \gamma).
    \end{aligned}
\end{equation}

Similarly, the oscillation probability from the iteration $t+1$ to $t+2$ is:
\begin{equation}
    \begin{aligned}
    \left.P({\bf b}^{{\bf w}^{t+1}} \neq {\bf b}^{{\bf w}^{t+2}})\right|_{{\bf b}^{{\bf w}^{t+1}} = 1} \leq P(\frac{\partial \mathcal{L}}{\partial \hat{\bf w}^{t+1}} \ge \gamma).
    \end{aligned}
\end{equation}

Thus, the sequential oscillation probability from the iteration $t$ to $t+2$ is:
\begin{equation}\label{eq:sequential}
    \begin{aligned}
    &P(({\bf b}^{{\bf w}^{t+1}} \neq {\bf b}^{{\bf w}^{t+2}}) \cap ({\bf b}^{{\bf w}^{t+1}} \neq {\bf b}^{{\bf w}^{t+2}})) |_{{\bf b}^{{\bf w}^{t}} = -1} \\
    &\leq P\big((\frac{\partial \mathcal{L}}{\partial \hat{\bf w}^t} \le  - \gamma) \cap (\frac{\partial \mathcal{L}}{\partial \hat{\bf w}^{t+1}} \ge  \gamma)\big),
    \end{aligned}
\end{equation}
which denotes that the weight oscillation happens only if the magnitudes of $\frac{\partial \mathcal{L}}{\partial \hat{\bf w}^{t}}$ and $\frac{\partial \mathcal{L}}{\partial \hat{\bf w}^{t+1}}$ are both larger than $\gamma$. 
\textbf{As a result, its attached factor $\gamma$ can be considered as a parameter used to control the occurrence of the weight oscillation}. 

However, if the conditions in Eq.\,(\ref{eq:sequential}) are met, with  Eq.\,(\ref{eq:chain_rule}) concluded, the gradient of $\hat{\bf w}^{t+1}$ is formulated as:
\begin{equation}
    \begin{aligned}
    \frac{\partial \mathcal{L}}{\partial \hat{\bf w}^{t+1}} &= \frac{\partial \mathcal{L}}{\partial \hat{\bf w}^t} - \eta \frac{\partial^2 \mathcal{L}}{\partial ({\hat{\bf w}^{t}})^2} \ge\gamma, \\
    \eta \frac{\partial^2 \mathcal{L}}{\partial ({\hat{\bf w}^{t}})^2} &\le \frac{\partial \mathcal{L}}{\partial \hat{\bf w}^{t}} -  \gamma \le -2 \gamma.
    \end{aligned}
\end{equation}

Note that $\eta$ and $\gamma$ are two positive variables, thus the second-order gradient $\frac{\partial^2 \mathcal{L}}{\partial {(\hat{\bf w}^{t}})^2} < 0$ holds always. Consequently, $\mathcal{L}(\hat{\bf w}^{t+1})$ can only be a local maxima, instead of a minima, which raises a contradiction with convergence in the training process. Such a contradiction indicates that the training algorithm will be convergent until no oscillation occurs, due to the additional term $\mathcal{S}(\alpha, {\bf w})$. Therefore, we completes our proof.
%
%The additional term $\mathcal{S}(\alpha, {\bf w})$ will self-stabilize the training process, when the weight flip phenomenon happens. Thus, our proposition is proved.
\end{myproof}

Our proposition and proof reveal that the balanced parameter $\gamma$ is actually a ``threshold''.
A very small ``threshold'' fails to mitigate the frequent oscillation effectively while a too large one suppresses the necessary sign inversion and hinders the gradient descent process.
To solve this, we devise the learning rule of $\gamma$ as:
\begin{equation}
\gamma^{n, t+1}_i = 
%\mathop{\rm max}_{1 \leq j\leq C_{in}\times K \times K}(|\frac{\partial \mathcal{L}}{\partial \hat{\bf w}^{n,t}_{i,j}}|)
\frac{1}{M^n}\|{\bf b}^{{\bf w}^{n,t}_i}\circledast {\bf b}^{{\bf w}^{n,t+1}_i} - {\bf 1}\|_0\cdot\mathop{\max}_{1 \leq j\leq M^n}(|\frac{\partial \mathcal{L}}{\partial \hat{\bf w}^{n,t}_{i,j}}|),
\label{eq:stabilizer}
\end{equation}
where the first item $\frac{1}{M^n}\|{\bf b}^{{\bf w}^{n,t}_i}\circledast {\bf b}^{{\bf w}^{n,t+1}_i} - {\bf 1}\|_0$ denotes the proportion of weights with sign changed. The second item $\mathop{\max}_{1 \leq j\leq M^n}(|\frac{\partial \mathcal{L}}{\partial \hat{\bf w}^{n,t}_{i,j}}|)$ is derived from Eq.\,(\ref{eq:sequential}), denoting the gradient with the greatest magnitude of the $t$-th iteration. In this way, we suppress the frequent weight oscillation by a resilient gradient. 

%Based on the proposition we improve the learning and stability of BNNs by a simple learnable scaling factor.
We further optimize the scaling factor as:
\begin{equation}
\delta_{{\alpha}^n_i}=\frac{\partial \mathcal{L}}{\partial {\alpha}^n_i}+\frac{\partial \mathcal{L}_R}{\partial {\alpha}^n_i}.
\label{27}
\end{equation}

The gradient derived from softmax loss can be easily calculated according to back propagation. Based on Eq.\,(\ref{10}), it is easy to derive:
\begin{equation}
\frac{\partial \mathcal{L}_R}{\partial {\alpha}^n_i}=\gamma^n_i{({\bf w}^n_i-{\alpha}^n_i {\bf b}^{{\bf w}^n_i})\circledast {\bf b}^{{\bf w}^n_i}}.
\label{29}
\end{equation}

\section{Experiments}
Our ReBNNs are evaluated first on image classification and object detection tasks for visual recognition. Then, we evaluate ReBNN on the GLUE~\cite{wang2018glue} benchmark with diverse NLP tasks. 
In this section, we first introduce the implementation details of ReBNN. Then we validate the effectiveness of the balanced parameter in the ablation study. Finally, we compare our method with state-of-the-art BNNs on various tasks to demonstrate the superiority of ReBNNs. 

\begin{figure*}[t]
    \centering
    \includegraphics[scale=.57]{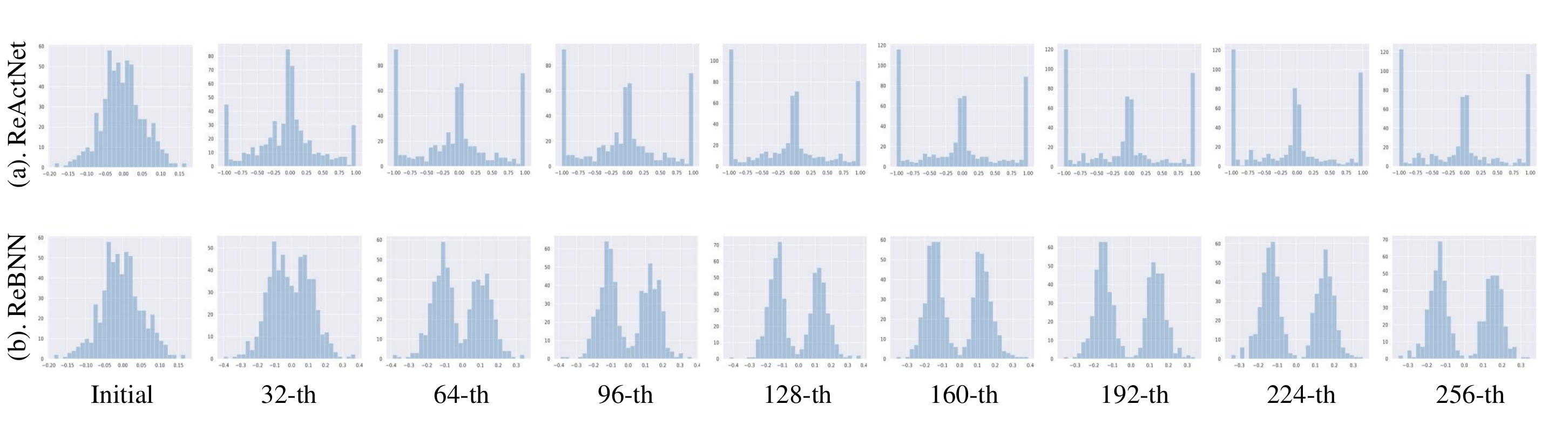}
    \caption{The evolution of latent weight distribution of (a) ReActNet and (b) ReBNN. We select the first channel of the first binary convolution layer to show the evolution.
    The model is initialized from the first stage training with W32A1 following~\cite{liu2020reactnet}. We plot the distribution every 32 epochs.}
    \label{fig:dist}
\end{figure*}

\subsection{Datasets and Implementation Details}
\label{secdata}
\textbf{Datasets:}
% CIFAR10~\cite{krizhevsky2009learning} is a raw image classification dataset containing a training set of $50,000$ and a testing set of $10,000$ $32\times 32$ color images across the following ten classes: airplanes, automobiles, birds, cats, deers, dogs, frogs, horses, ships, and trucks.
For its huge scope and diversity, the ImageNet object classification dataset~\cite{imagenet15} is more demanding, which has 1000 classes, 1.2 million training photos, and 50k validation images.

The COCO dataset includes images from 80 different categories. All our experiments on COCO dataset are conducted on the COCO {\tt 2014}~\cite{coco2014} object detection track in the training stage, which contains the combination of 80k images from the COCO {\tt train2014} and 35k images sampled from COCO {\tt val2014}, {\em i.e.}, COCO {\tt trainval35k}. Then we test our method on the remaining 5k images from the COCO {\tt minival}. We report the average precision (AP) for IoUs$\in$[0.5: 0.05: 0.95], designated as mAP@[.5,.95], using COCO's standard evaluation metric. For further analyzing our method, we also report AP$_{\rm 50}$, AP$_{\rm 75}$, AP$_s$, AP$_m$, and AP$_l$.

The GLUE benchmark contains multiple natural language understanding tasks. We follow~\cite{wang2018glue} to evaluate the performance: Matthews correlation for CoLA, Spearman correlation for STS-B, and accuracy for the rest tasks: RTE, MRPC, SST-2, QQP, MNLI-m (matched), and MNLI-mm (mismatched). Also, for machine reading comprehension on SQuAD,
we report the EM (exact match) and F1 score. 

\noindent\textbf{Implementation Details:}
PyTorch~\cite{paszke2017automatic} is used to implement ReBNN. We run the experiments on 4 NVIDIA Tesla A100 GPUs with $80$ GB memory. Following~\cite{liu2018bi}, we retain weights in the first layer, shortcut, and last layer in the networks as the real-valued.

For the image classification task, ResNets~\cite{he2016deep} are employed as the backbone networks to build our ReBNNs. We offer two implementation setups for fair comparison. First, we use \textbf{one-stage training} on ResNets, with SGD as the optimization algorithm and a momentum of 0.9, and a weight decay of $1e\!-\!{4}$ following~\cite{xu2021recu}. $\eta$ is set to 0.1. The learning rates are optimized by the annealing cosine learning rate schedule. The number of epochs is set as 200. Then, we employ \textbf{two-stage training} following~\cite{liu2020reactnet}. Each stage counts 256 epochs. In this implementation, Adam is selected as the optimizer. And the network is supervised by a real-valued ResNet-34 teacher. The weight decay is set as 0 following~\cite{liu2020reactnet}. The learning rates $\eta$ is set as $1e\!-\!3$ and annealed to 0 by linear descent.

For objection detection, we use the Faster-RCNN~\cite{ren2016faster} and SSD~\cite{liu2016ssd}, which are based on ResNet-18~\cite{he2016deep} and VGG-16~\cite{Simonyan15} backbone, respectively. We fine-tune the detector on the dataset for object detection. For SSD and
\begin{table}[h]
	\caption{We compare different calculation method of $\gamma$, including constant varying from $0$ to $1e\!-\!2$ and gradient-based calculation.}
	\centering
\setlength{\tabcolsep}{2mm}{
    %\caption{We compare different calculation method of $\gamma$, including constant varying from $0$ to $1e\!-\!2$ and gradient-based calculation.}
\begin{tabular}{ccc}
\hline
Value of $\gamma$                                                                                    & Top-1         & Top-5         \\ \hline
$0$                                                                                                  & 65.8          & 86.3          \\
$1e\!-\!5$                                                                                               & 66.2          & 86.7          \\
$1e\!-\!4$                                                                                               & 66.4          & 86.7          \\
$1e\!-\!3$                                                                                               & 66.3          & 86.8          \\
$1e\!-\!2$                                                                                               & 65.9          & 86.5          \\ \cdashline{1-3}
$\mathop{\max}_{1\leq j\leq M^n}(|\frac{\partial \mathcal{L}}{\partial \hat{\bf w}^{n,t}_{i,j}}|)$ & 66.3          & 86.2          \\
\textbf{Eq.~(\ref{eq:stabilizer})}                                                                               & \textbf{66.9} & \textbf{87.1} \\ \hline
    \end{tabular}}
    %\vspace{-1mm}
	\label{gamma}
\end{table}
Faster-RCNN, the batch size is set to 16 and 8, respectively, with applying SGD optimizer. $\eta$ is equal to 0.008. We use the same structure and training settings as BiDet~\cite{wang2020bidet} on the SSD framework. The input resolution is $1000\times600$ for Faster-RCNN and $300\times300$ for SSD, respectively.

For the natural language processing task, we conduct experiments based on BERT$_{\rm BASE}$ \cite{devlin2018bert} (with 12 hidden layers) architecture following BiBERT~\cite{qin2022bibert}, respectively. The detailed training setups are the same as BiBERT. We extend the ReBNN to multi-layer perceptrons (MLPs) and use the Bi-Attention following BiBERT~\cite{qin2022bibert}.

\subsection{Ablation study}
Since no extra hyper-parameter is introduced, we first evaluate the different calculation of $\gamma$. Then we show how our ReBNN achieves a resilient training process. In the ablation study, we use the ResNet-18 backbone initialized from the first stage training with W32A1 following~\cite{liu2020reactnet}.

\noindent{\bf Calculation of $\gamma$:} We compare the different calculations of  $\gamma$ in this part. As shown in Tab.~\ref{gamma}, the performances increase first and then decrease when increasing the value of constant $\gamma$. Considering that the gradient magnitude varies layer-wise and channel-wise, a subtle $\gamma$ can hardly be manually set as a global value. We further compare the gradient-based calculation. To avoid extreme values, we set the upper bound as $2e-4$ and lower bound as $1e-5$.
As shown in the bottom lines, we first use $\mathop{\max}_{1\leq j\leq M^n}(|\frac{\partial \mathcal{L}}{\partial \hat{\bf w}^{n,t}_{i,j}}|)$, the maximum intra-channel gradient 
\begin{figure}[t]
	\centering
	\includegraphics[scale=.4]{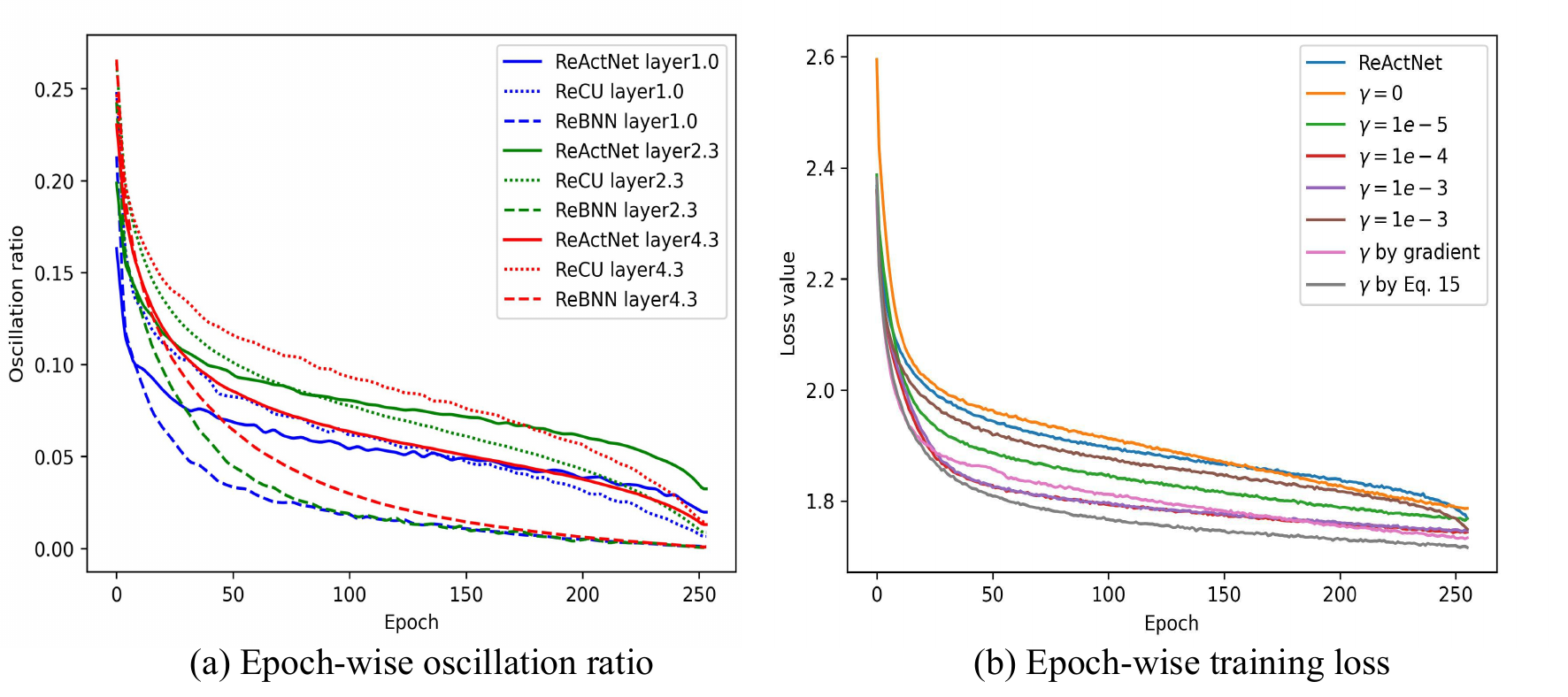}
	\figcaption{(a) The epoch-wise weight oscillation ratio of ReActNet (solid), ReCU (dotted) and ReBNN (dashed). (b) Comparing the loss curves of ReActNet and our ReBNN with different calculation of $\gamma$.}
	\label{flip_loss}
\end{figure}
of last iteration, which shows a similar performance compared with the constant $1e\!-\!4$. 
This indicates that only using the maximum intra-channel gradient may also suppress necessary sign flip, thus hindering the training. 
Inspired by this, we use Eq.~(\ref{eq:stabilizer}) to calculate $\gamma$ and improve the performance by 0.6\%, showing that considering the weight oscillation proportion allows the necessary sign flip and leads a more effective training. 
We also show the training loss curves in Fig.~\ref{flip_loss}(b). As plotted, the curves of $\mathcal{L}$ almost demonstrate the degrees of training sufficiency. Thus we draw the conclusion that ReBNN with $\gamma$ calculated by Eq.~(\ref{eq:stabilizer}) achieves the lowest training loss as well as an efficient training process. 
\begin{table*}[]
\caption{A performance comparison with SOTAs on ImageNet with different training strategies. \#Bits denotes the bit width of weights and activations. We report the Top-1 (\%) and Top-5 (\%) accuracy performances. ReBNN$_1$ and ReBNN$_2$ denote our ReBNN learned with one-stage and two-stage training.}
\renewcommand\arraystretch{0.86}
\centering
\small
\setlength{\tabcolsep}{3mm}{\begin{tabular}{ccccccc}
\hline
Network                     & Method      & \#Bits               & Size$_{\rm (MB)}$        & OPs$_{(10^8)}$          & Top-1 & Top-5 \\ \hline
\multirow{11}{*}{ResNet-18} & Real-valued & 32-32                & 46.76                  & 18.21                 & 69.6  & 89.2  \\ \cdashline{2-7} 
                            & BNN          & \multirow{7}{*}{1-1} & \multirow{7}{*}{4.15}  & \multirow{7}{*}{1.63} & 42.2  & 67.1  \\
                            & XNOR-Net    &                      &                        &                       & 51.2  & 73.2  \\
                            & Bi-Real Net &                      &                        &                       & 56.4  & 79.5  \\
                            & RBNN       &                      &                        &                       & 59.6  & 81.6  \\
                            & ReCU       &                      &                        &                       & 61.0  & 82.6  \\
                            & \textbf{ReBNN$_1$}      &                      &                        &                       & \textbf{61.6}  & \textbf{83.4}  \\ \cdashline{2-7} 
                            & ReActNet    & \multirow{4}{*}{1-1} & \multirow{4}{*}{4.15}  & \multirow{4}{*}{1.63} & 65.9  & 86.2  \\
                            & FDA-BNN       &                      &                        &                       & 65.8  & 86.4  \\
                            & ReCU       &                      &                        &                       & 66.4  & 86.5  \\
                            & \textbf{ReBNN$_2$}      &                      &                        &                       & \textbf{66.9}  & \textbf{87.1}  \\ \hline
\multirow{7}{*}{ResNet-34}  & Real-valued  & 32-32                & 87.19                  & 36.74                 & 73.3  & 91.3  \\ \cdashline{2-7} 
                            & Bi-Real Net & \multirow{4}{*}{1-1}                  & \multirow{4}{*}{5.41} & \multirow{4}{*}{1.93} & 62.2  & 83.9  \\
                            & RBNN       &                      &                        &                       & 63.1  & 84.4  \\
                            & ReCU       &                      &                        &                       & 65.1  & 85.8  \\
                            & \textbf{ReBNN$_1$}      &  &                        &                       & \textbf{65.8}  & \textbf{86.2}  \\ \cdashline{2-7} 
                            & ReActNet    & \multirow{2}{*}{1-1} & \multirow{2}{*}{5.41} & \multirow{2}{*}{1.93} & 69.3  & 88.6  \\
                            & \textbf{ReBNN$_2$}      &                      &                        &                       & \textbf{69.9}  & \textbf{88.9}  \\ \hline
\end{tabular}}
\label{imagenet}
\end{table*}

\begin{table*}[h]
\small
\caption{Comparison of mAP@[.5, .95](\%), AP (\%) with different IoU threshold and AP for objects in various sizes with state-of-the-art binarized object detectors on COCO {\tt minival}. \#Bits denotes the bit width of weights and activations}
\centering
{
\begin{tabular}{cccccccccccc}
\hline
Framework                     & Backbone                   & Method          & \#Bits               & Size$_{\rm (MB)}$      & OPs$_{\rm (G)}$         & \begin{tabular}[c]{@{}c@{}}mAP\\ @{[}.5, .95{]}\end{tabular} & AP$_{50}$     & AP$_{75}$     & AP$_s$       & AP$_m$        & AP$_1$        \\ \hline
\multirow{6}{*}{Faster-RCNN} & \multirow{6}{*}{ResNet-18} & Real-valued     & 32-32                & 47.48                  & 434.39                & 26.0                      & 44.8          & 27.2          & 10.0         & 28.9          & 39.7          \\ \cdashline{3-12} 
                              &                            & DeRoFa-Net     & 4-4                  & 6.73                   & 55.90                 & 22.9                      & 38.6          & 23.7          & 8.0          & 24.9          & 36.3          \\ \cdashline{3-12} 
                              &                            & XNOR-Net
                             & \multirow{4}{*}{1-1} & \multirow{4}{*}{2.39}  & \multirow{4}{*}{8.58} & 10.4                      & 21.6          & 8.8           & 2.7          & 11.8          & 15.9          \\
                              &                            & Bi-Real Net   &                      &                        &                       & 14.4                      & 29.0          & 13.4          & 3.7          & 15.4          & 24.1          \\
                              &                            & BiDet        &                      &                        &                       & 15.7                      & 31.0          & 14.4          & 4.9          & 16.7          & 25.4          \\
                              &                            & \textbf{ReBNN} &                      &                        &                       & \textbf{19.6}             & \textbf{37.6} & \textbf{20.4} & \textbf{7.0} & \textbf{20.1} & \textbf{33.1} \\ \hline
\multirow{6}{*}{SSD}          & \multirow{6}{*}{VGG-16}    & Real-valued    & 32-32                & 105.16                 & 31.44                 & 23.2                      & 41.2          & 23.4          & 5.3          & 23.2          & 39.6          \\ \cdashline{3-12} 
                              &                            & DoReFa-Net     & 4-4                  & 29.58                  & 6.67                  & 19.5                      & 35.0          & 19.6          & 5.1          & 20.5          & 32.8          \\ \cdashline{3-12} 
                              &                            & XNOR-Net       & \multirow{4}{*}{1-1} & \multirow{4}{*}{21.88} & \multirow{4}{*}{2.13} & 8.1                       & 19.5          & 5.6           & 2.6          & 8.3           & 13.3          \\
                              &                            & Bi-Real Net    &                      &                        &                       & 11.2                      & 26.0          & 8.3           & 3.1          & 12.0          & 18.3          \\
                              &                            & BiDet     &                      &                        &                       & 13.2                      & 28.3          & 10.5          & \textbf{5.1} & 14.3          & 20.5          \\
                              &                            & \textbf{ReBNN} &                      &                        &                       & \textbf{18.1}             & \textbf{33.9} & \textbf{17.5} & 4.2          & \textbf{17.9} & \textbf{25.9} \\ \hline
\end{tabular}}
\label{COCO}
\end{table*}
Note that the loss may not be minimal at each training iteration, but our method is just a reasonable version of gradient descent algorithms by nature, which can be used to solve the optimization problem as general. We empirically prove ReBNN's capability of mitigating the weight oscillation, leading to a better convergence.

\noindent{\bf Resilient training process:} We first show the evolution of the latent weight distribution is this section. We plot the distribution of the first channel of the first binary convolution layer per 32 epochs in Fig.~\ref{fig:dist}. As seen, our ReBNN can efficiently redistribute the BNNs towards resilience. Conventional ReActNet~\cite{liu2020reactnet} possesses a tri-model distribution, which is unstable due to the scaling factor with large magnitudes. In contrast, our ReBNN is constrained by the balanced parameter $\gamma$ during training, thus leading to a resilient bi-modal distribution with fewer weights centering around zero. We also plot the ratios of sequential weight oscillation of ReBNN and ReActNet for the 1-st, 8-th, and 16-th binary convolution layers of ResNet-18. As shown in Fig.~\ref{flip_loss}(a), the dashed lines gain much lower magnitudes than the solid (ReActNet) and dotted (ReCU \cite{xu2021recu}) lines with the same color, validating the effectiveness of our ReBNN in suppressing the consecutive weight oscillation. Besides, we can also find that the sequential weight oscillation ratios of ReBNN are gradually decreased to 0 as the training converges. 
%The supplementary material includes more analyses regarding the layer-wise weight oscillation.

\subsection{Image classification}
We first show the experimental results on ImageNet with ResNet-18 and ResNet-34~\cite{he2016deep} backbones in Tab.~\ref{imagenet}. We compare ReBNN with BNN~\cite{courbariaux2015binaryconnect}, XNOR-Net~\cite{rastegari2016xnor}, Bi-Real Net~\cite{liu2018bi}, RBNN~\cite{lin2020rotated}, and ReCU~\cite{xu2021recu} for the one-stage training strategy. For the two-stage training strategy, we compare with ReActNet~\cite{liu2020reactnet}, FDA-BNN~\cite{xu2021learning}, and ReCU~\cite{xu2021recu}.

\begin{table*}[t]
\centering
\small
\caption{Comparison of BERT quantization methods without data augmentation.  \#Bits denotes the bit width of weights, word embedding, and activations. “Avg.” denotes the average results.}
\setlength{\tabcolsep}{1mm}{\begin{tabular}{ccccccccccccc}
\hline
Method                 & \#Bits                 & Size$_{\rm (MB)}$             & OPs$_{\rm (G)}$                & MNLI-m/mm            & QQP                  & QNLI                 & SST-2                & CoLA                     & STS-B                & MRPC                 & RTE                  & Avg.                              \\ \hline
Real-valued           & 32-32-32               & 418                   & 22.5                  & 84.9/85.5            & 91.4                 & 92.1                 & 93.2                 & 59.7                     & 90.1                 & 86.3                 & 72.2                 & 83.9                              \\ \cdashline{1-13}
Q-BERT              & 2-8-8                  & 43.0                  & 6.5                   & 76.6/77.0            & -                    & -                    & 84.6                 & -                        & -                    & 68.3                 & 52.7                 & -                                 \\
Q2BERT               & 2-8-8                  & 43.0                  & 6.5                   & 47.2/47.3            & 67.0                 & 61.3                 & 80.6                 & 0                        & 4.4                  & 68.4                 & 52.7                 & 47.7                              \\
TernaryBERT           & 2-2-2                  & 28.0                  & 1.5                   & 40.3/40.0            & 63.1                 & 50.0                 & 80.7                 & 0                        & 12.4                 & 68.3                 & 54.5                 & 45.5                              \\ \cdashline{1-13}
BinaryBERT       & \multirow{3}{*}{1-1-1} & \multirow{3}{*}{16.5} & \multirow{3}{*}{0.4}  & 35.6/35.3            & 66.2                 & 51.5                 & 53.2                 & 0                        & 6.1                  & 68.3                 & 52.7                 & 41.0                              \\
BiBERT          &                        &                       &                       & 66.1/67.5            & 84.8                 & 72.6                 & 88.7                 & 25.4                     & 33.6                 & 72.5                 & \textbf{57.4}        & 63.2                              \\
\textbf{ReBNN}                 &                        &                       &                       & \textbf{69.9/71.3}   & \textbf{85.2}        & \textbf{79.2}        & \textbf{89.3}        & \textbf{28.8}            & \textbf{38.7}        & \textbf{72.6}        & 56.9                 & \textbf{65.8} \\ \hline
\end{tabular}}
\label{nlp}
\end{table*}

\begin{table*}[t]
\centering
\small
\caption{Comparing ReBNN with real-valued models on hardware (single thread).}
\setlength{\tabcolsep}{2mm}{\begin{tabular}{ccccccc}
\hline
Network                     & Method      & W/A   & Size$_{\rm (MB)}$ & Memory Saving             & Latency$_{\rm (ms)}$ & Acceleration              \\ \hline
\multirow{2}{*}{ResNet-18}  & Real-valued & 32/32 & 46.76      & -                         & 583.1       & -                         \\\cdashline{2-7}
                            & ReBNN       & 1/1   & 4.15       & 11.26$\times$ &    67.5      & 8.64$\times$ \\ \cdashline{1-7}
\multirow{2}{*}{ResNet-34}  & Real-valued & 32/32 & 87.19& - & 1025.6 &-                        \\\cdashline{2-7}
& ReBNN       & 1/1   & 5.41       & 16.12$\times$ & 113.6         & 9.03$\times$  \\ \hline
% \multirow{2}{*}{ReActNet-A} & Real-valued & 32/32 &      117.7     &             -              &     608.1         &         -                  \\
%                             & ReBNN       & 1/1   &     7.8      &         15.1\times                  &       144.5       & 4.2\times                          \\ \hline
\end{tabular}}
\label{hardware}
%%\vspace{-6mm}
\end{table*}

We evaluate our ReBNN with one-stage training following~\cite{xu2021recu}. ReBNN outperforms all of the compared binary models in both Top-1 and Top-5 accuracy, as shown in Tab.~\ref{imagenet}. ReBNN-based ResNet-18 respectively achieves 61.6\% and 83.4\% in Top-1 and Top-5 accuracy, with 0.6\% increases over state-of-the-art RECU. ReBNN further outperforms all compared methods with ResNet-34 backbone, achieving 65.8\% Top-1 accuracy. We further evaluate our ReBNN with two-stage training following~\cite{liu2020reactnet}, where ReBNN surpasses ReActNet by 1.0\%/0.6\% Top-1 accuracy with ResNet-18/34 backbones, respectively.

In this paper, we use memory usage and OPs following~\cite{liu2018bi} in comparison to other tasks for further reference. As shown in Tab.~\ref{imagenet}, ReBNN theoretically accelerates ResNet-18/34 by 11.17$\times$ and 19.04$\times$, which is significant for real-time applications. 
%More analyses regarding the training cost are shown in the supplementary material.

\subsection{Object detection}
Because of its size and diversity, the COCO dataset presents a great challenge for object detection. On COCO, the proposed ReBNN is compared against state-of-the-art 1-bit neural networks such as XNOR-Net~\cite{rastegari2016xnor}, Bi-Real Net~\cite{liu2018bi}, and BiDet~\cite{wang2020bidet}. We present the performance of the 4-bit DoReFa-Net~\cite{zhou2016dorefa} for reference. 

As shown in Tab.~\ref{COCO}, compared with state-of-the-art XNOR-Net, Bi-Real Net, and BiDet, our method improves the mAP@[.5,.95] by 9.2\%, 5.2\%, and 3.9\% using the Faster-RCNN framework with the ResNet-18 backbone. Moreover, on other APs with different IoU thresholds, our ReBNN clearly beats others. Compared to DoReFa-Net, a quantized neural network with 4-bit weights and activations, our ReBNN obtains only 3.3\% lower mAP. Our method yields a 1-bit detector with a performance of only 6.1\% mAP lower than the best-performing real-valued counterpart (19.6\% {\em vs.} 26.0\%). Similarly, using the SSD300 framework with the VGG-16 backbone, our method achieves 18.1\% mAP@[.5,.95], outperforming XNOR-Net, Bi-Real Net, and BiDet by 10.0 \%, 6.9\%, and 4.9\% mAP, respectively. Our ReBNN also achieves highly efficient models by theoretically accelerating Faster-RCNN and SSD by 50.62$\times$ and 14.76$\times$.

\subsection{Natural language processing}
In Tab.~\ref{nlp}, we show experiments on the BERT$_{\rm BASE}$ architecture and the GLUE benchmark without data
augmentation following BiBERT~\cite{qin2022bibert}.  Experiments show that outperforms other methods on the development set of GLUE benchmark, including TernaryBERT~\cite{zhang2020ternarybert}, BinaryBERT~\cite{bai2020binarybert}, Q-BERT~\cite{shen2020q}, Q2BERT~\cite{shen2020q}, and BiBERT~\cite{qin2022bibert}.  Our ReBNN surpasses existing methods on BERT$_{\rm BASE}$ architecture by a clear margin in the average accuracy. For example, our ReBNN surpasses BiBERT by 6.6\% accuracy on QNLI dataset, which is significant for the natural language processing task. We observe our ReBNN brings improvements on 7 out of total 8 datasets, thus leading to a 2.6\% average accuracy improvement. 
Our ReBNN also achieves highly efficient models by theoretically accelerating the BERT$_{\rm BASE}$ architecture by 56.25$\times$.

\subsection{Deployment Efficiency}
\label{deploy}

We implement the 1-bit models achieved by our ReBNN on ODROID C4, which has a 2.016 GHz 64-bit quad-core ARM Cortex-A55. With evaluating its real speed in practice, the efficiency of our ReBNN is proved when deployed into real-world mobile devices. 
We leverage the SIMD instruction SSHL on ARM NEON to make the inference framework BOLT \cite{feng2021bolt} compatible with ReBNN.
%To make inference framework BOLT \cite{feng2021bolt} compatible with ReBNN, we use the SIMD instruction SSHL on ARM NEON. 
We compare ReBNN to the real-valued backbones in Tab. \ref{hardware}. We can see that ReBNN's inference speed is substantially faster with the highly efficient BOLT framework. For example, the acceleration rate achieves about 8.64$\times$ on ResNet-18, which is slightly lower than the theoretical acceleration rate. For ResNet-34 backbone, ReBNN can achieve 9.03$\times$ acceleration rate with BOLT framework on hardware, which is significant for the computer vision on real-world edge devices.

\section{Conclusion}
In this paper, we analyze the influence of frequent weight oscillation in binary neural networks and proposed a Resilient Binary Neural Network (ReBNN) to provide resilient gradients for latent weights updating.
Our method specifically proposes to parameterize the scaling factor and introduces a weighted reconstruction loss to build an adaptive training objective. We further manifest that the balanced parameter can serve as an indicator to reflect the frequency of the weight oscillation during back propagation. 
Our ReBNN reaches resilience by learning the balanced parameter, leading to a great reduction of weight oscillation. ReBNN shows strong generalization to gain impressive performance on various tasks such as image classification, object detection, and natural language processing tasks, demonstrating the superiority of the proposed method over state-of-the-art BNNs.
% Use \bibliography{yourbibfile} instead or the References section will not appear in your paper

\section{Acknowledgment}
This work was supported by National Natural Science Foundation of China under Grant 62076016, 62206272, 62141604, 61972016, and 62032016, Beijing Natural Science Foundation L223024.

\bibliography{aaai23}

\end{document}